\pdfoutput=1
\documentclass[11pt]{article}

\usepackage[preprint]{acl}

\usepackage{times}
\usepackage{latexsym}
\usepackage{amsfonts}
\usepackage{authblk} 
\usepackage{amsmath}
\usepackage{multirow}
\usepackage{longtable}
\usepackage{subcaption}  
\usepackage{enumitem}
\usepackage{booktabs} 

\usepackage[T1]{fontenc}

\usepackage[utf8]{inputenc}

\usepackage{microtype}

\usepackage{inconsolata}

\usepackage{graphicx}
\title{CycleDistill: Bootstrapping Machine Translation using LLMs\\ with Cyclical Distillation}

\author{
  \textbf{Deepon Halder}$^{1,4}$\hspace{0.2cm}
  \textbf{Thanmay Jayakumar}$^{1,2}$\hspace{0.2cm}
  \textbf{Raj Dabre}$^{1,2,3}$\thanks{Corresponding Author: \href{mailto:raj.dabre@cse.iitm.ac.in}{raj.dabre@cse.iitm.ac.in}}
  \\
  $^{1}$Nilekani Centre at AI4Bharat \quad
  $^{2}$Indian Institute of Technology, Madras \\
  $^{3}$Indian Institute of Technology, Bombay \\
  $^{4}$Indian Institute of Engineering, Science and Technology, Shibpur
}

\begin{document}
    \maketitle
    \begin{abstract}
        Large language models (LLMs), despite their ability to perform few-shot machine translation (MT), often lag behind dedicated MT systems trained on parallel corpora, which are crucial for high quality machine translation (MT). However, parallel corpora are often scarce or non-existent for low-resource languages. In this paper, we propose CycleDistill, a bootstrapping approach leveraging LLMs and few-shot translation to obtain high-quality MT systems. CycleDistill involves iteratively generating synthetic parallel corpora from monolingual corpora via zero- or few-shot MT, which is then used to fine-tune the model that was used for generating said data for MT. CycleDistill does not need parallel corpora beyond 1 to 4 few-shot examples, and in our experiments focusing on three Indian languages, by relying solely on monolingual corpora, it can achieve high-quality machine translation, improving upon a few-shot baseline model by \textbf{ 20-30 chrF points} on average in the first iteration. We also study the effect of leveraging softmax activations during the distillation process and observe mild improvements in translation quality. We publicly release the source code associated with this work\footnote{Code : \href{https://github.com/deeps73/CycleDistill}{Github}}.
    \end{abstract}

    \section{Introduction}

Machine translation (MT) for low-resource languages poses persistent challenges due to the limited availability of bilingual corpora and the linguistic variation these languages exhibit. Although large language models (LLMs) can perform translation with minimal supervision, their effectiveness in low-resource settings is typically inferior to systems trained with substantial parallel data~\cite{koehn2017six, gu2018universal}.

This paper introduces \textit{CycleDistill}, a resource-efficient framework for improving translation quality without requiring extensive parallel data. The approach begins with a small set of example translations and utilizes LLMs to generate synthetic parallel corpora from monolingual text. These corpora are then used to iteratively fine-tune the translation model, enabling progressive performance gains with each cycle.

The framework incorporates two key techniques. First, \textit{Iterative Synthetic Data Distillation} leverages repeated cycles of data generation and model training to enhance translation performance over time~\cite{kim2021improving}. Second, \textit{Soft Distribution-Preserving Distillation} transfers detailed token-level probability distributions from teacher to student models, allowing for more comprehensive knowledge retention~\cite{tan2019multilingual}. Building on previous work in self-training~\cite{he2019revisiting}, sequence-level and soft-target knowledge distillation~\cite{kim2016sequence, hinton2015distilling}, \textit{CycleDistill} offers a practical and scalable solution for MT in low-resource scenarios.

\begin{figure}
    \centering
    \includegraphics[width=1\linewidth]{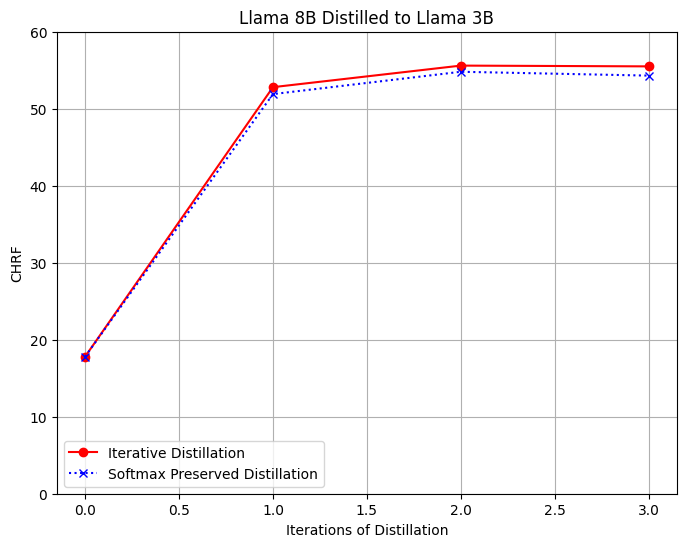}
    \caption{chrF scores over distillation cycles for LLaMA 8B \textrightarrow{} 3B using Iterative and Softmax-Preserved Distillation under a zero-shot Hindi setting. Marginal gains observed across iterations.}
    \label{fig:hindi_distill_cycles}
\end{figure}

The main contributions of this work are:

\begin{itemize}[noitemsep, topsep=0.5pt]
    \item We present \textit{CycleDistill}, a self-supervised MT framework that improves translation quality using only monolingual corpora and minimal supervision.
    \item We propose a token-level soft distillation strategy to facilitate richer and more effective learning from teacher models.
    \item We demonstrate that our method achieves substantial improvements of 20-30 chrF points over few-shot translation baselines, with consistent chrF score gains across three Indian low-resource languages.
\end{itemize}

\section{Related work} 
Low-resource machine translation (MT) remains a significant challenge due to the scarcity of parallel corpora and high linguistic diversity~\citep{koehn2017six,gu2018universal}. Knowledge distillation (KD) has become a popular approach for addressing these issues, transferring knowledge from large teacher models to smaller student models~\citep{hinton2015distilling}. Sequence-level KD~\citep{kim2016sequence} and iterative or self-training strategies~\citep{kim2021improving,furlanello2018born} have demonstrated improvements in low-resource and multilingual MT~\citep{tan2019multilingual}. Recent advances include continual KD, which sequentially distills knowledge from multiple existing models~\citep{zhang2023continual}, and encoder-aware KD for better transfer in compute-constrained and low-resource settings~\citep{velayuthan2025encoder}.

Back-translation and its iterative variants are also highly effective for low-resource MT, as they leverage monolingual data to generate synthetic parallel corpora~\citep{edunov2018understanding,hoang2018iterative}. These methods have shown strong gains in extremely low-resource and Indic language scenarios, especially when combined with transfer learning and data filtering~\citep{luo2020joint,tars2021extremely,ahmed2023iterative,krishnamurthy2024mtnlp}. 

While both KD and back-translation have advanced the field, their integration and comparative effectiveness, particularly in settings with minimal parallel supervision, remain active areas of research.
Our proposed \textbf{CycleDistill} framework is novel in that it bootstraps high-quality MT systems using only monolingual corpora and a handful of few-shot examples, without relying on large-scale parallel data. Unlike prior work, CycleDistill combines cyclical iterative synthetic data generation with token-level soft distribution-preserving distillation, enabling progressive model refinement and compression.
    
    \section{Methodology}
    
    This work aims to enhance low resource languages to English machine translation through the adoption of two iterative distillation strategies: cyclic synthetic data generation and an advanced distillation approach that preserves detailed token-level information, such as softmax distributions and subword structures. Our methodology is grounded in recent developments in knowledge distillation and self-training for neural machine translation~\cite{kim2016sequence, gou2021knowledge}.
    \begin{figure}
        \centering
        \includegraphics[width=1\linewidth]{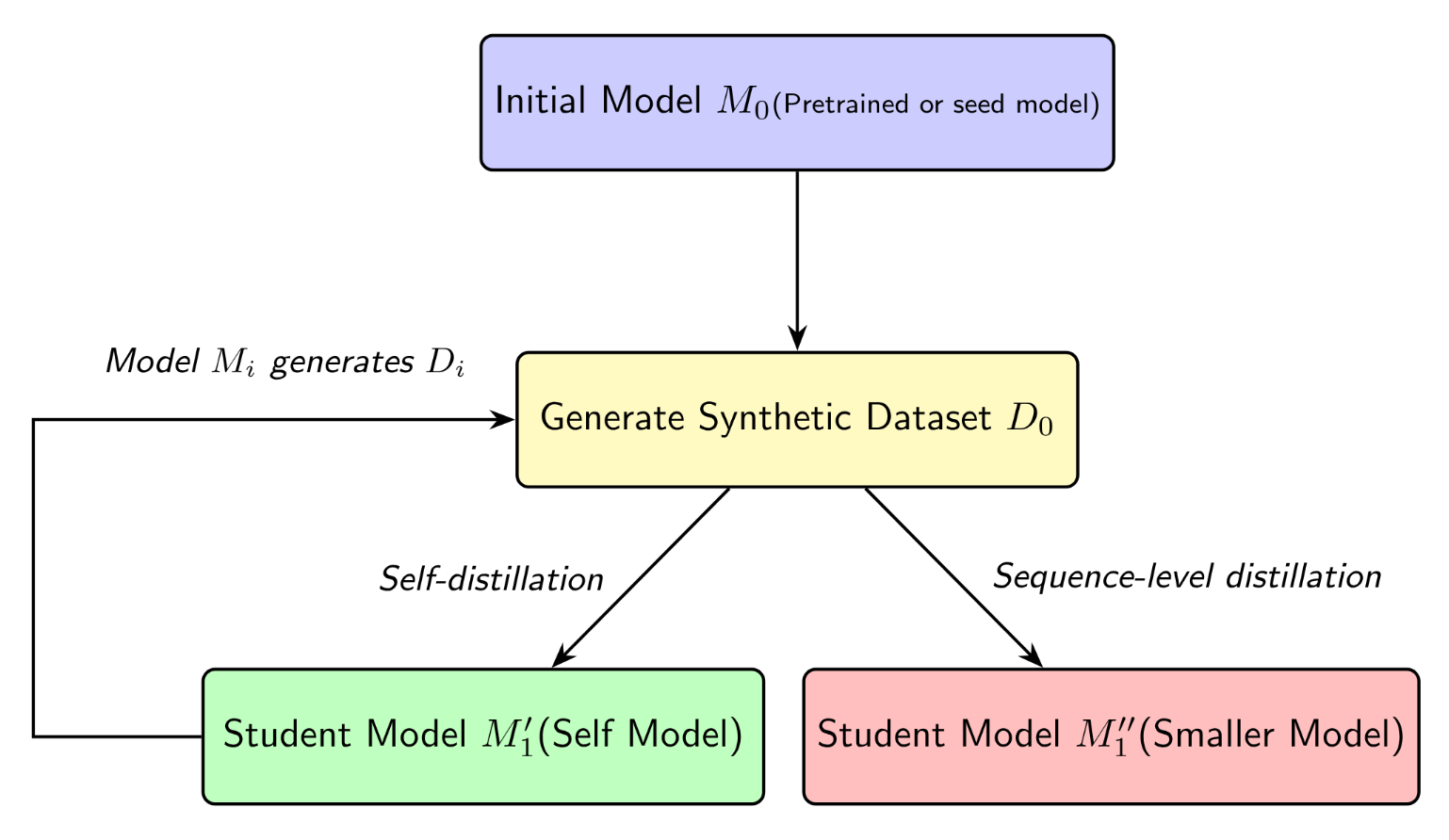}
        \caption{An overview of the CycleDistill framework,
which iteratively generates synthetic parallel data from
monolingual corpora and refines translation models
through cyclic self distillation.}
        \label{fig:framework_overview}
    \end{figure}

    \subsection{Iterative Synthetic Data Distillation}
    
    Our first approach enables the base translation model to iteratively improve by generating and learning from its own synthetic data. The procedure is as follows:
    
    \begin{itemize}
        \item \textbf{Base Model Initialization:} The process begins with a pretrained base translation model, denoted as \( M_0 \), which is capable of translating from an Indic language to English.
        
        \item \textbf{Synthetic Data Generation:} The model \( M_0 \) is employed to generate a synthetic dataset \( D_0 \) comprising translation pairs. This step is inspired by self-training methodologies that have demonstrated efficacy in low-resource scenarios~\cite{he2019revisiting}.
        
        \item \textbf{Self-Distillation:} Utilizing the generated synthetic data, knowledge distillation is performed in two ways:
        \begin{itemize}
            \item The same model architecture is further refined, resulting in an updated model \( M_1 \).
            \item Additionally, knowledge is distilled into a smaller student model, \( M_1' \), via sequence-level knowledge distillation, whereby the student learns from the teacher's generated translations~\cite{kim2016sequence}.
        \end{itemize}
        
        \item \textbf{Iterative Refinement:} This procedure is repeated for three cycles. In each iteration \( i \) (where \( i = 1, 2, 3 \)):
        \begin{itemize}
            \item The distilled model \( M_i \) (or \( M_i' \)) produces a new dataset \( D_i \) comprising additional translations.
            \item Subsequently, \( M_i \) is distilled into \( M_{i+1} \) and a new student model \( M_{i+1}' \).
        \end{itemize}
    \end{itemize}
    
    The underlying principle is that, by iteratively learning from its own outputs, the model can progressively improve its performance. This iterative process benefits both the primary and the student models, enhancing their generalization capabilities and, in certain cases, enabling model size reduction with minimal compromise in performance.
    
    \subsection{Soft Distribution-Preserving Distillation}
    
    The second strategy extends the distillation process by capturing more granular information from the teacher model:
    
    \begin{itemize}
        \item \textbf{Enhanced Data Extraction:} During synthetic translation generation, for each token position \( t \), we record:
        \begin{itemize}
            \item The top-\( k \) token predictions (\( \{y_1^{(t)}, \ldots, y_{k}^{(t)}\} \))~\cite{fan2018hierarchical}
            \item The corresponding softmax probabilities (\( \{p_1^{(t)}, \ldots, p_{k}^{(t)}\} \)), where \( \sum_{j=1}^{k} p_j^{(t)} \leq 1 \)
        \end{itemize}
        This comprehensive information set is motivated by the demonstrated effectiveness of soft-target distillation in capturing the teacher model's knowledge~\cite{hinton2015distilling}.
        \begin{figure}
            \centering
            \includegraphics[width=1.0\linewidth]{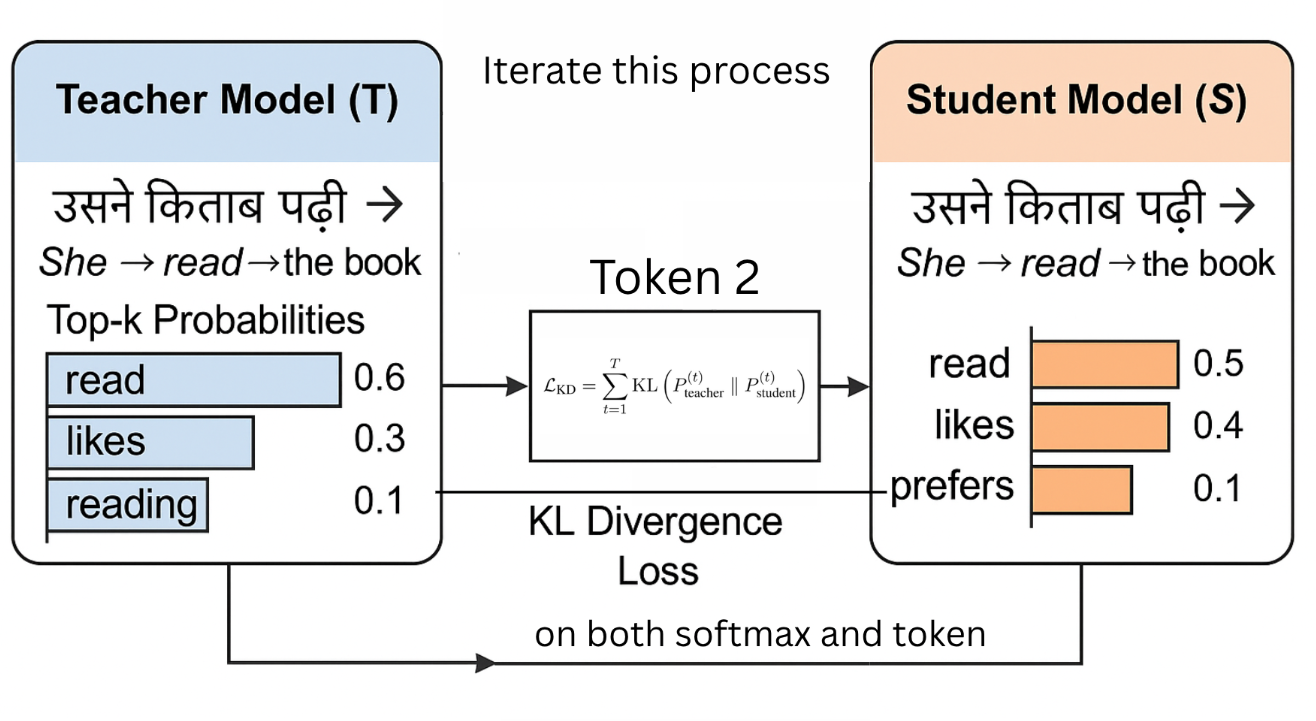}
            \caption{An Overview of the \textit{Soft Distribution Preserving Distillation}. Unlike standard distillation, this method preserves top-k token distributions at each position. The student model learns not only from final outputs but also from the richer probability landscape, encouraging finer-grained generalization.}
            \label{fig:soft_distill_overview}
        \end{figure}
        \item \textbf{Logit-Based Distillation:} The student model is trained to match not only the final output sequences but also the softmax distributions over the top-\( k \) tokens at each position. This is achieved by minimizing the Kullback-Leibler (KL) divergence \citep{kullback1951information} loss:
        \[
        \mathcal{L}_{\text{KD}} = \sum_{t=1}^{T} \mathrm{KL} \left( P^{(t)}_{\text{teacher}} \parallel P^{(t)}_{\text{student}} \right)
        \]
        where \( T \) denotes the sequence length, and \( P^{(t)} \) represents the softmax distributions. This approach enables the student model to more accurately approximate the teacher's behavior, as suggested in prior research~\cite{hinton2015distilling, mukherjee2021distilling}.
        
        \item \textbf{Iterative Distillation:} This process is also conducted over three iterations. In each cycle, the student from the previous round assumes the role of the new teacher, and a fresh synthetic dataset is generated, ensuring the transfer of rich token-level distributions.
    \end{itemize}

    \section{Experiments}
    
    This section outlines the experimental framework designed to investigate the efficacy of iterative knowledge distillation in enhancing machine translation quality. Our approach involves distilling knowledge from larger language models into smaller counterparts, followed by comprehensive performance evaluation across multiple metrics and languages.
    
    \subsection{Models and Languages}
    
    Our study employs four language models of varying sizes from the LLaMA~\cite{meta2024llama3} and Gemma~\cite{google2024gemma2} families:
    
\begin{itemize}[noitemsep, topsep=2pt]
    \item \textbf{Gemma 2 9B} (\( G_{9B} \))
    \item \textbf{Gemma 2 2B} (\( G_{2B} \))
    \item \textbf{LLaMA 3.1 8B} (\( L_{8B} \))
    \item \textbf{LLaMA 3.2 3B} (\( L_{3B} \))
\end{itemize}
    
    Each larger model undergoes distillation to produce both a refined same-size model and a compressed smaller model, adhering to established Sequence Distillation principles~\cite{kim2016sequence}. Our evaluation encompasses three Indic languages:

    \begin{itemize}[noitemsep, topsep=2pt]
        \item \textbf{Hindi} (\(HIN \))
        \item \textbf{Bengali} (\(BEN \))
        \item \textbf{Malayalam} (\(MAL \))
    \end{itemize}
    \subsection{Distillation Process}
    
    For a given teacher model \( T \), distillation is performed to produce two student models:
    \begin{itemize}[noitemsep, topsep=2pt]
        \item Same-size student (\( S_{\text{same}} \leftarrow T \))
        \item Smaller student (\( S_{\text{small}} \leftarrow T \))
    \end{itemize}
    
    The distillation relationships are formally expressed as:
    \[
    G_{9B} \rightarrow \{ G'_{9B}, G_{2B} \}, \quad L_{8B} \rightarrow \{ L'_{8B}, L_{3B} \}
    \]
    where the refined large models (\( G'_{9B}, L'_{8B} \)) are subsequently utilized for synthetic data generation. We select \( k = 20 \) after empirical evaluation of the teacher models’ output distributions revealed that the probability mass beyond the 20 highest-scoring tokens is negligible. We perform the experiments only upto three iterations (\( n = 3 \)).
    This limit was set because we observed that the performance gains stabilized after the third iteration. Further iterations yielded negligible improvements, indicating that the models were approaching a performance plateau, making additional computational cycles inefficient.
    
    \subsection{Training Data}
    
    Models are fine-tuned using the \textbf{BPCC seed corpus} ~\cite{gala2023indictrans2}, a parallel Indic-to-English dataset. Consistent with established practices in low-resource translation research~\cite{kunchukuttan2023indicnlp}, we randomly sample 20,000 sentence pairs for training and distillation. We use a fixed prompt format for all of the language and model pair, discussed in Figure \ref{lol}. 
    
    \subsection{Synthetic Data Generation}
    
    Following each distillation iteration, the most recent large model generates synthetic English translations for the original 20,000 source sentences. This synthetic data generation process is repeated for three complete cycles to enable progressive model refinement.
    \subsection{Prompt Used}
\begin{figure}
    \centering
    \includegraphics[width=1\linewidth]{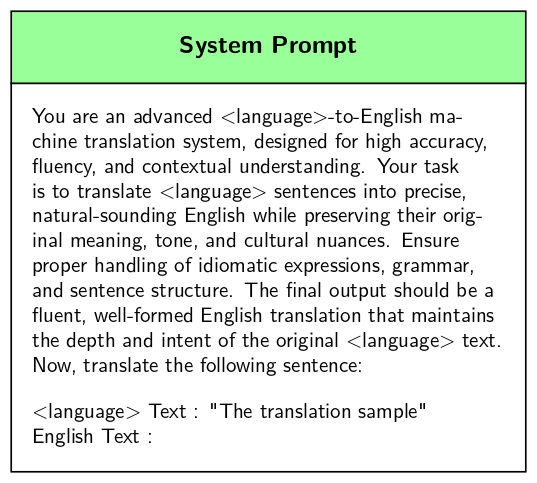}
    \caption{Example of the general prompt used for the translation task.}
    \label{lol}
\end{figure}

The prompt utilized for the translation task described in Section 4.3 is shown in Figure \ref{lol}.

In 1-shot and 4-shot settings, example translation pairs are inserted into the middle section of the prompt prior to the final instruction.
    
    \subsection{Evaluation}
    
    Model performance is assessed using the \textbf{IN22 Gen corpus}~\cite{gala2023indictrans2}, the standard evaluation benchmark coupled with the BPCC seed corpus. The translation quality is quantified through chrF scores~\cite{popovic2015chrf}. This metric provides standardized measurement of n-gram translation accuracy, aligning with current best practices in machine translation evaluation.
\begin{table*}[!t]
\centering
\scriptsize
\setlength{\tabcolsep}{4pt} 
\renewcommand{\arraystretch}{1.2} 
\begin{tabular}{ll rrr rrr rrr} 
\toprule
\multirow{2}{*}{\textbf{Model}} & \multirow{2}{*}{\textbf{Iter}} & \multicolumn{3}{c}{\textbf{chrF (0-shot)}} & \multicolumn{3}{c}{\textbf{chrF (1-shot)}} & \multicolumn{3}{c}{\textbf{chrF (4-shot)}} \\
\cmidrule(lr){3-5} \cmidrule(lr){6-8} \cmidrule(lr){9-11}
& & \textbf{BEN} & \textbf{HIN} & \textbf{MAL} & \textbf{BEN} & \textbf{HIN} & \textbf{MAL} & \textbf{BEN} & \textbf{HIN} & \textbf{MAL} \\
\midrule
\multirow{7}{*}{$G_{9B}$} & Base & 41.4 & 47.9 & 39.9 & 42.7 & 49.2 & 38.8 & 24.2 & 44.6 & 14.5 \\\cline{2-11}
& $DD_1$ & 61.1 & 64.4 & 60.2 & 60.8 & 64.2 & 60.0 & 53.1 & 63.8 & 37.0 \\
& $SD_1$ & 60.9 & 64.7 & 60.4 & 60.1 & 64.5 & 57.9 & 49.3 & 63.7 & 18.2 \\
\cmidrule(lr){2-11}
& $DD_2$ & 61.4 & 64.5 & 60.7 & 60.5 & 64.6 & 60.2 & 52.4 & 63.7 & 37.2 \\
& $SD_2$ & 60.5 & 64.7 & 60.7 & 64.8 & 64.9 & 59.1 & 49.3 & 64.3 & 32.9 \\
\cmidrule(lr){2-11}
& $DD_3$ & 61.0 & 60.4 & 61.1 & 60.6 & 59.0 & 60.4 & 52.8 & 57.7 & 37.8 \\
& $SD_3$ & 61.4 & 64.4 & 61.0 & 60.9 & 63.3 & 58.4 & 45.0 & 64.1 & 48.1 \\
\midrule
\multirow{7}{*}{$L_{8B}$} & Base & 29.2 & 33.6 & 22.8 & 26.6 & 36.0 & 8.5 & 13.5 & 24.1 & 14.0 \\\cline{2-11}
& $DD_1$ & 44.9 & 29.8 & 42.6 & 39.6 & 26.8 & 17.6 & 16.7 & 18.9 & 17.4 \\
& $SD_1$ & 42.1 & 40.3 & 40.6 & 32.0 & 39.6 & 21.2 & 16.7 & 29.3 & 17.4 \\
\cmidrule(lr){2-11}
& $DD_2$ & 48.3 & 50.3 & 46.2 & 42.0 & 55.5 & 26.4 & 16.5 & 51.1 & 17.4 \\
& $SD_2$ & 46.2 & 54.1 & 44.5 & 38.3 & 39.4 & 23.5 & 15.1 & 33.4 & 17.4 \\
\cmidrule(lr){2-11}
& $DD_3$ & 38.9 & 37.3 & 17.8 & 30.0 & 27.6 & 15.0 & 18.3 & 21.0 & 17.4 \\
& $SD_3$ & 38.9 & 50.8 & 38.0 & 38.7 & 40.7 & 22.3 & 17.0 & 27.3 & 17.4 \\
\midrule
\multirow{7}{*}{$L_{3B}$} & Base & 24.2 & 14.5 & 2.9 & 18.4 & 17.8 & 5.0 & 13.4 & 14.5 & 14.0 \\\cline{2-11}
& $DD_1$ & 46.0 & 52.7 & 38.9 & 39.3 & 52.8 & 27.4 & 27.0 & 36.3 & 17.4 \\
& $SD_1$ & 49.4 & 53.1 & 33.5 & 37.5 & 51.9 & 18.2 & 17.2 & 34.5 & 17.3 \\
\cmidrule(lr){2-11}
& $DD_2$ & 34.3 & 55.0 & 37.5 & 28.0 & 55.6 & 24.5 & 12.8 & 42.7 & 17.3 \\
& $SD_2$ & 52.3 & 54.4 & 29.4 & 39.3 & 54.8 & 17.5 & 16.6 & 44.4 & 17.2 \\
\cmidrule(lr){2-11}
& $DD_3$ & 26.1 & 55.1 & 27.1 & 16.4 & 55.5 & 18.7 & 13.4 & 42.6 & 17.4 \\
& $SD_3$ & 45.2 & 53.9 & 25.3 & 37.5 & 54.3 & 17.4 & 13.5 & 42.8 & 17.3 \\
\midrule
\multirow{7}{*}{$G_{2B}$} & Base & 24.6 & 28.8 & 23.8 & 28.7 & 33.4 & 27.8 & 19.0 & 31.2 & 13.4 \\\cline{2-11}
& $DD_1$ & 50.9 & 58.4 & 48.3 & 50.3 & 58.7 & 46.6 & 27.7 & 54.1 & 25.4 \\
& $SD_1$ & 40.1 & 58.3 & 48.2 & 58.3 & 56.9 & 47.1 & 23.8 & 55.5 & 23.0 \\
\cmidrule(lr){2-11}
& $DD_2$ & 50.0 & 58.1 & 48.2 & 50.1 & 58.4 & 47.1 & 29.0 & 53.8 & 25.8 \\
& $SD_2$ & 43.0 & 58.4 & 49.0 & 48.8 & 58.1 & 47.4 & 28.6 & 51.2 & 21.4 \\
\cmidrule(lr){2-11}
& $DD_3$ & 49.9 & 57.8 & 47.4 & 49.4 & 57.2 & 46.9 & 34.9 & 54.9 & 25.3 \\
& $SD_3$ & 49.1 & 56.8 & 48.5 & 45.4 & 56.8 & 47.0 & 32.8 & 53.3 & 21.0 \\
\midrule
\textbf{Average} & & \textbf{44.4} & \textbf{51.5} & \textbf{40.9} & \textbf{39.8} & \textbf{49.6} & \textbf{31.0} & \textbf{26.8} & \textbf{42.5} & \textbf{21.6} \\
\bottomrule
\end{tabular}
\caption{chrF scores for all models and methods across three languages and shot settings, with column averages.}
\label{tab:chrF_scores_wide_col_avg}
\end{table*}
    
    \section{Results and Analyses}
    We first describe our main results on CycleDistill (iterative self distillation) and then analyze its various effects.
    \subsection{Main Results}
    \noindent \textbf{Zero-Shot Setting}
    We observe a consistent performance trend across iterations of distillation. The first iteration results in a substantial performance increase. The second and third iteration usually have similar values with the first iteration, but we notice a small increase of 1-2\% of chrF with each iteration.
    
    This pattern holds true for both \textit{iterative distillation} and \textit{soft distribution-preserving distillation}, with no significant differences observed between the two. However there are some notable results --
    \begin{itemize}[itemsep=1pt, topsep=2pt]
        \item For the Gemma 2B model with Bengali and the LLaMA 3B model with Malayalam, iterative distillation outperforms soft distribution-preserving distillation.
        \item In contrast, for the LLaMA 8B model with Hindi and the LLaMA 3B model with Bengali, soft distribution-preserving distillation demonstrates superior performance compared to iterative distillation.
    \end{itemize}
    
    \noindent \textbf{One-Shot Setting}
    The one-shot setting yields the best overall performance, with the highest chrF scores observed exclusively in this configuration. The performance trend across iterations closely resembles that of the zero-shot setting. We observe some crossover between the two distillation methods, where one approach outperforms the other depending on the iteration count. Notable observations include:
    \begin{itemize}[itemsep=1pt, topsep=2pt]
        \item For the LLaMA 3B model on the Malayalam dataset, iterative distillation surpasses soft distribution-preserving distillation in performance.
        \item Conversely, for the LLaMA 3B model on the Bengali dataset, soft distribution-preserving distillation outperforms iterative distillation.
    \end{itemize}

    \noindent \textbf{Four-Shot Setting}
Performance declines slightly in the four-shot setting compared to earlier configurations, though iteration-wise differences remain minimal. Both iterative and soft distribution-preserving distillation exhibit similar gradual improvements and overall trends. This drop is primarily attributed to reduced contextual clarity due to increased input length, the four-shot prompt is approximately 60\% longer than the one-shot, placing greater demands on the model’s context window. Maintaining coherence across multiple examples becomes harder as prompts grow longer. The degradation is more pronounced in linguistically complex languages, suggesting that context dilution disproportionately affects grammatically rich targets. These results highlight the need to balance shot count and context efficiency in multilingual distillation, especially under limited model capacities.

\subsection{Impact of Language Morphology on chrF}
To further investigate the observed decline in 4-shot performance, particularly for morphologically rich languages, we visualize language-specific sensitivity to increasing shot settings. As shown in Table 1, we find a notable and steeper decline from 1-shot to 4-shot for Bengali and Malayalam, compared to Hindi, which supports the hypothesis that context dilution disproportionately impacts morphologically complex languages.

\begin{table*}[!t]
\centering
\scriptsize
\setlength{\tabcolsep}{5pt} 
\renewcommand{\arraystretch}{1.2} 
\begin{tabular}{ll rrr rrr}
\toprule
\multirow{2}{*}{\textbf{Model}} & \multirow{2}{*}{\textbf{Iter}} & \multicolumn{3}{c}{\textbf{Nepali (Devanagari Script)}} & \multicolumn{3}{c}{\textbf{Manipuri (Meitei Script)}} \\
\cmidrule(lr){3-5} \cmidrule(lr){6-8}
& & \textbf{0-shot} & \textbf{1-shot} & \textbf{4-shot} & \textbf{0-shot} & \textbf{1-shot} & \textbf{4-shot} \\
\midrule
\multirow{7}{*}{$L_{8B}$} & Base & 12.47  & 13.95  & --  & 16.88 & 17.45 & 17.45 \\\cline{2-8}
& $DD_1$ & 38.59 & 38.08 & --  & 18.51 & 17.74 & 17.75 \\
& $SD_1$ & 54.44  & 36.19 & -- & 16.97 & 17.61 & 17.43    \\
\cmidrule(lr){2-8}
& $DD_2$ & 35.23  & 30.45 & -- & 18.52 & 17.02 & 17.17 \\
& $SD_2$ & 54.31 & 35.19 & -- & 18.84 & 17.82 & 18.08 \\
\cmidrule(lr){2-8}
& $DD_3$ & 33.24  & 20.38 & -- & 17.87 & 15.97 & 15.98 \\
& $SD_3$ & 54.74 & 34.35 & -- & 18.04 & 16.98 & 16.93 \\
\midrule
\multirow{7}{*}{$L_{3B}$} & Base & 17.16  & 17.15 & -- & 17.13 & 17.44 & 17.45 \\\cline{2-8}
& $DD_1$ & 48.55  & 48.75  & -- & 18.58 & 16.82 & 17.41 \\
& $SD_1$ & 47.31  & 25.51 & -- & 18.70 & 16.77 & 16.81 \\
\cmidrule(lr){2-8}
& $DD_2$ & 40.48 & 38.23  & --  & 17.88 & 14.74 & 14.57 \\
& $SD_2$ & 47.31 & 25.67 & -- & 17.35 & 15.11 & 14.81 \\
\cmidrule(lr){2-8}
& $DD_3$ & 41.15  & 39.34 & --  & 17.49 & 15.73 & 15.59 \\
& $SD_3$ & 47.08 & 31.11 & -- & 17.08 & 13.64 & 13.47 \\
\bottomrule
\end{tabular}
\caption{chrF scores for Nepali (Devanagari script) and Manipuri (Meitei script) over the Llama model family.}
\label{tab:nepali_manipuri_combined}
\end{table*}

\subsection{Effectiveness in Extremely Low Resource Languages }
\noindent \textbf{Study on Nepali }
To assess the robustness and generalizability of our proposed method in low-resource or moderately known language settings, we conducted experiments using Meta’s LLaMA 3.1 8B and LLaMA 3.2 3B models. We selected Nepali, written in the Devanagari script, as the target language. This language has partial representation in the model's pretraining corpus, which means the models possess a basic understanding of it and are capable of generating reasonable outputs, although it is not extensively covered. Despite this limited exposure, the models were able to produce useful distillation data. When we applied our method, we observed consistent improvements over baseline methods, as shown in Table 2. These results suggest that our method remains effective even when the target language has minimal presence in the training data. This demonstrates the potential of our approach to enhance performance in low-resource and cross-lingual generalization scenarios.

\noindent \textbf{Study on Manipuri }
The investigation included preliminary experiments on the Manipuri (Meitei script) to English translation task, utilizing several prominent large language models, specifically GPT-4, LLaMA 3.1 8B, and Gemma 2 9B. These models were evaluated for their ability to generate synthetic distillation data, which is the first step for the proposed CycleDistill framework.

Results indicated a significant limitation: none of the evaluated models were capable of producing usable distillation data for Manipuri. This suggests that the process is inherently constrained in environments where the base large language model cannot effectively perform few-shot translation for the target low-resource language.
Further detailed experiments were conducted on Manipuri (Meitei script) using the LLaMA 3.1 8B and LLaMA 3.2 3B models within the iterative distillation framework. As presented in Table 2, these results consistently showed no improvement in chrF scores across successive iterations. 

\subsection{Further Analyses}

\paragraph{Teacher Quality vs. Student Gain\\}

To examine the correlation between teacher model performance and student gains within our CycleDistill framework\begin{figure}
    \centering
    \includegraphics[width=1\linewidth, height=0.6\textheight, keepaspectratio]{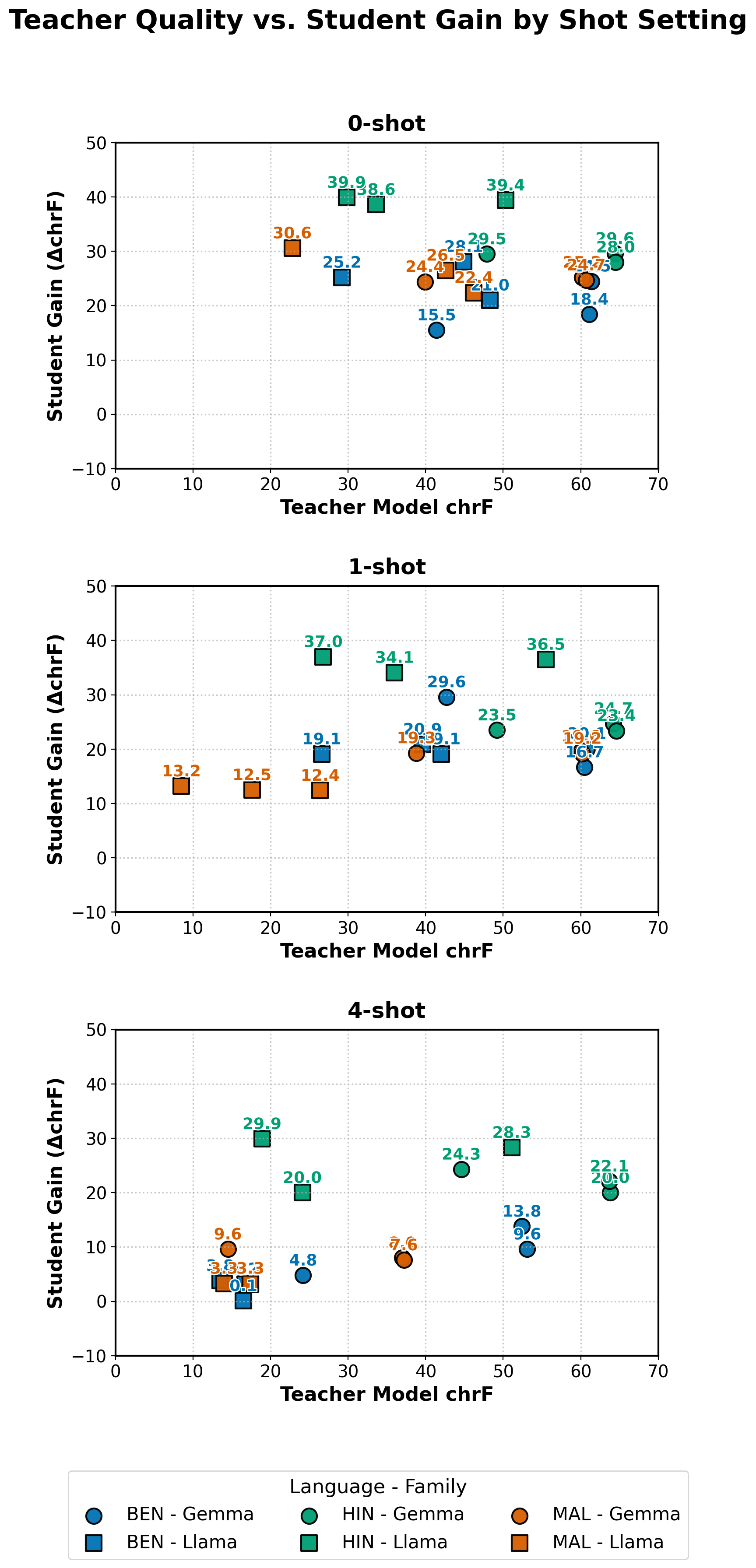}
    \caption{Scatter plot illustrating the relationship between teacher model performance and student model gain across zero-shot, one-shot, and four-shot settings in the CycleDistill framework.}
    \label{lmaoo}
\end{figure}, we analyzed the relevant data as depicted in Figure \ref{lmaoo}, where the 
x-axis indicates teacher performance (measured by the chrF score of models such as $G'^*_{9B}$ or $L'^*_{8B}$ when generating synthetic data), and the y-axis represents student gain ($\Delta$chrF, denoting the improvement over the baseline, e.g., $\text{chrF}^*_{G^*_{2B}\text{distilled}} - \text{chrF}^*_{G^*_{2B}\text{base}}$).

Our analysis reveals that this relationship varies by shot setting. In zero-shot, a positive correlation holds, with higher teacher scores driving greater gains, validating distillation's reliance on data quality in example-free scenarios. In one-shot, correlation vanishes, as a single example anchors learning, making gains independent of teacher quality. In four-shot, gains are suppressed overall, due to context dilution and error propagation in longer prompts, positioning one-shot as the optimal for effective distillation.

\paragraph{Error Propagation and Recovery\\}

A key limitation observed during our experiments is the susceptibility of the iterative framework to error propagation. Specifically, if an error such as the use of incorrectly generated or misaligned synthetic data is introduced at any iteration (for example, the second cycle), it can lead to a substantial degradation in performance, with declines of up to 30 to 40 chrF points observed in certain settings. These errors are compounded across subsequent iterations, as the model continues to self-distill based on flawed data, making recovery increasingly difficult. However, we also find that corrective interventions such as fine-tuning with accurately generated synthetic data can effectively mitigate such errors in subsequent iterations. This underscores the importance of early detection and correction of distillation errors, as well as the need for robust validation mechanisms during each cycle to prevent error amplification.

\paragraph{Performance of CycleDistill over Model Families\\}

\begin{figure}
    \centering
    \includegraphics[width=1\linewidth]{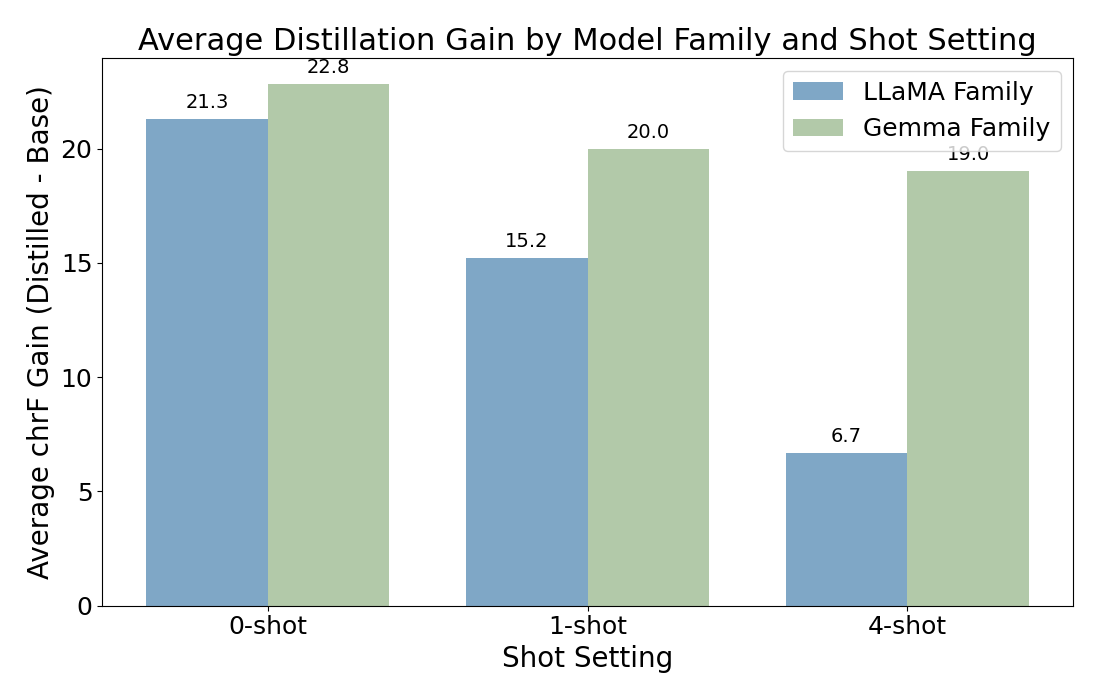}

    \caption{chrF gains for Gemma and LLaMA across shot settings.}
    \label{fig:performance_comparison}
\end{figure}

A key finding is the divergence in performance between LLaMA and Gemma models under CycleDistill, as shown in Figure~\ref{fig:performance_comparison}. Gemma exhibits superior, robust learning, as compared to LLaMA.

These results emphasize that the choice of base model architecture critically influences the stability and effectiveness of iterative distillation strategies.

\paragraph{Efficiency of Knowledge Absorption across Model Families\\}

The analysis of knowledge absorption rates reveals that the LLaMA 3B model exhibits a significantly higher efficiency in learning from its teacher compared to the Gemma 2B model. Specifically, the average absorption rate for LLaMA 3B is 1.190, while Gemma 2B achieves 0.628. This metric is defined as
\[
\text{Absorption Rate} = \frac{\text{Student Peak Gain}}{\text{Teacher Base Score}}
\]
where Student Peak Gain is the maximum chrF improvement over the student's base score across distillation iterations and Teacher Base Score is the teacher's initial chrF score, is averaged across nine evaluation conditions (three languages and three shot settings). Although the Gemma family demonstrates superior absolute chrF scores, supported by a stronger teacher (Gemma 9B), the LLaMA 3B's higher absorption rate suggests it is a more efficient learner, particularly beneficial in resource-constrained distillation scenarios.

\section{Conclusion}
    This work presents \textit{CycleDistill}, a structured and data-efficient framework for enhancing machine translation from low-resource languages to English. By leveraging iterative synthetic data generation and token-level soft distillation, CycleDistill improves translation performance without reliance on large-scale parallel corpora. Experimental results across multiple low-resource Indian languages confirm consistent gains in chrF scores, demonstrating the effectiveness of the approach under varying linguistic and architectural conditions.

The integration of iterative self-distillation with soft distribution-based learning reveals complementary benefits, though performance improvements taper beyond the second iteration, and translation quality remains sensitive to error accumulation, particularly in morphologically rich languages and limited supervision settings. Nevertheless, \textit{CycleDistill} enables both model refinement and compression without relying on large-scale parallel corpora, making it an efficient and scalable solution for low-resource MT and a meaningful contribution to multilingual NLP research.

    \section{Limitations}
    
    Despite the effectiveness of CycleDistill in enhancing translation performance through iterative and soft distribution-preserving distillation, the approach exhibits several notable limitations. Firstly, empirical results demonstrate diminishing marginal improvements beyond the second iteration, with performance frequently plateauing or deteriorating by the third cycle. Secondly, the method relies on synthetic data generated by teacher models, which may introduce compounding translation errors over successive iterations due to self-reinforcement effects. Thirdly, in few-shot scenarios, particularly involving morphologically rich languages such as Malayalam and Bengali, the system suffers significant performance degradation, up to 30 chrF points, largely attributable to increased prompt lengths and consequent loss of contextual coherence. Finally, the current evaluation is limited to three Indic languages and specific model families (Gemma and LLaMA), thereby restricting the generalizability of the findings to other language pairs and model architectures.
    
\section{Acknowledgements}
We would like to express our sincere gratitude to Dr. Anoop Kunchukuttan for his unwavering support, invaluable guidance, and insightful contributions throughout the course of this work. We also extend our thanks to EkStep Foundation and Nilekani Philanthropies for their generous grant, which has supported research efforts at AI4Bharat.

\bibliography{acl_latex}
\clearpage
\appendix

\section{Appendix A : Visualization of Effects of our Methods over Shots}
\label{sec:vis}
    This appendix provides a set of visualizations that illustrate the impact of the proposed methods under varying shot settings. Figures 7-11 demonstrate how performance characteristics evolve as the number of shots increases, thereby offering a more detailed understanding of the underlying behavior and effectiveness of our approach.

    \clearpage
\begin{figure*}
    \centering
    \includegraphics[width=\linewidth]{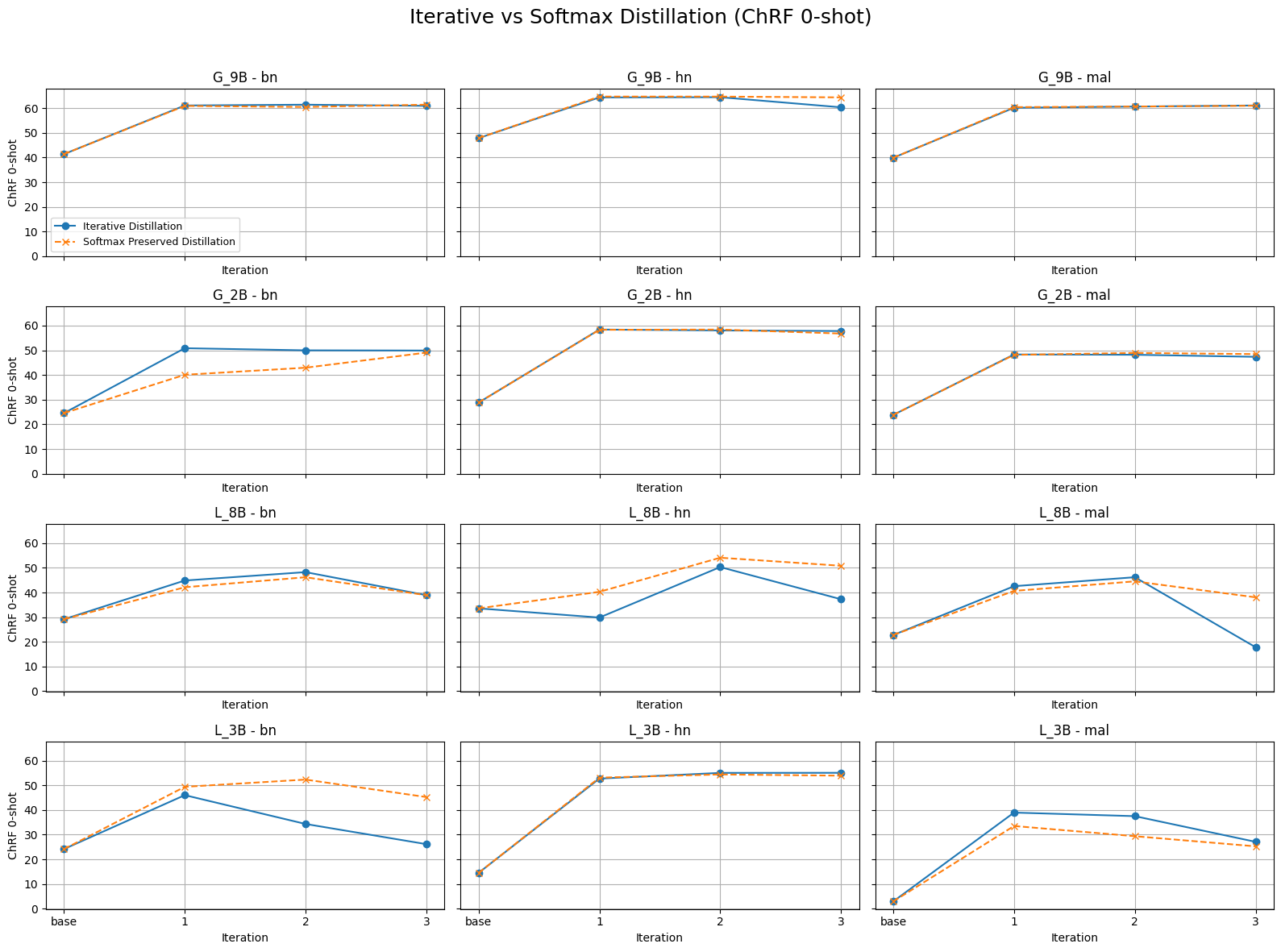}
    \caption{Comparison of the methods at 0-shot setting}
    \label{fig:vis_0shot}
\end{figure*}
\begin{figure*}
    \centering
    \includegraphics[width=\linewidth]{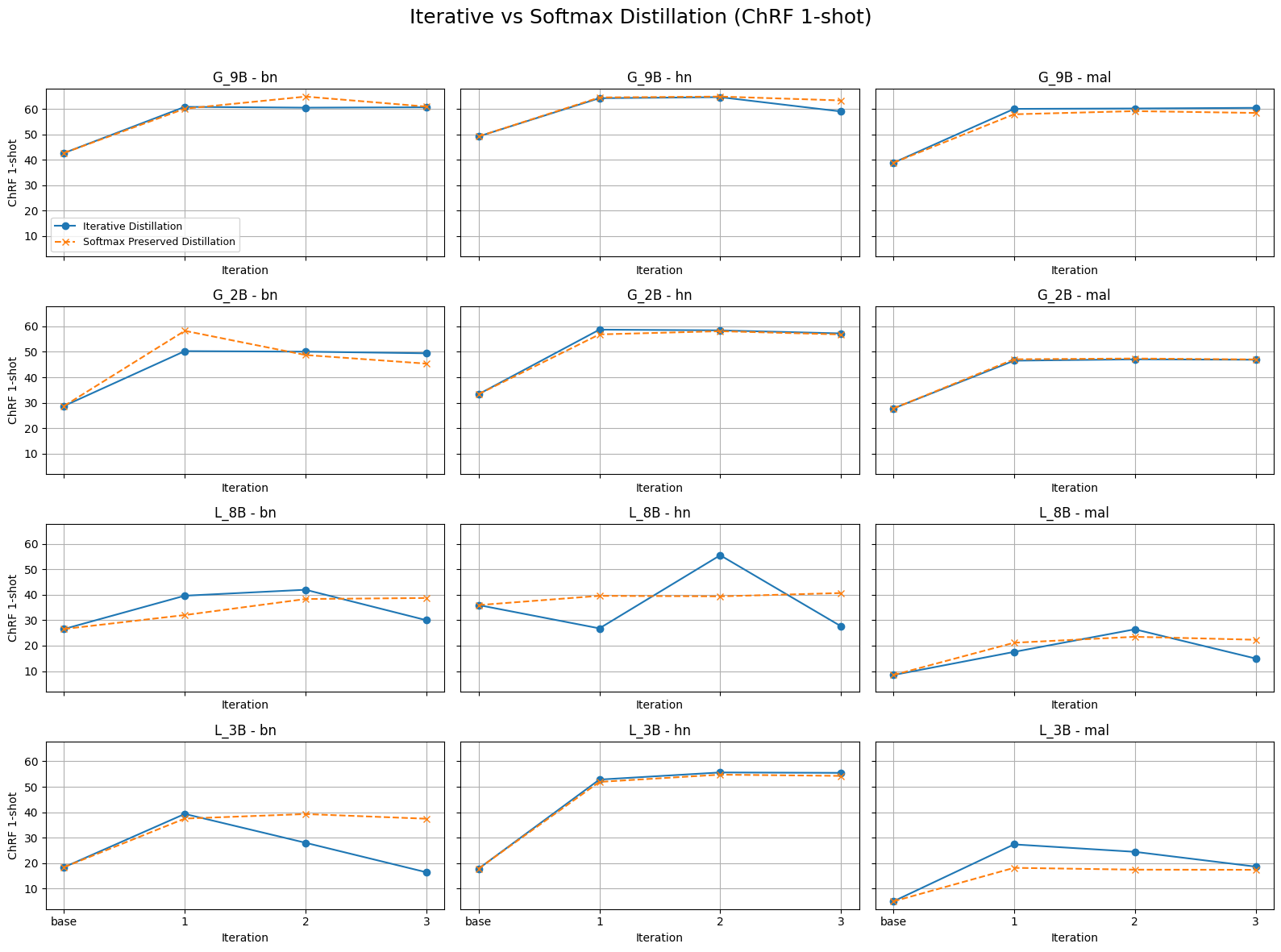}
    \caption{Comparison of the methods at 1-shot setting}
    \label{fig:vis_1shot}
\end{figure*}
\clearpage
\begin{figure*}
    \centering
    \includegraphics[width=\linewidth]{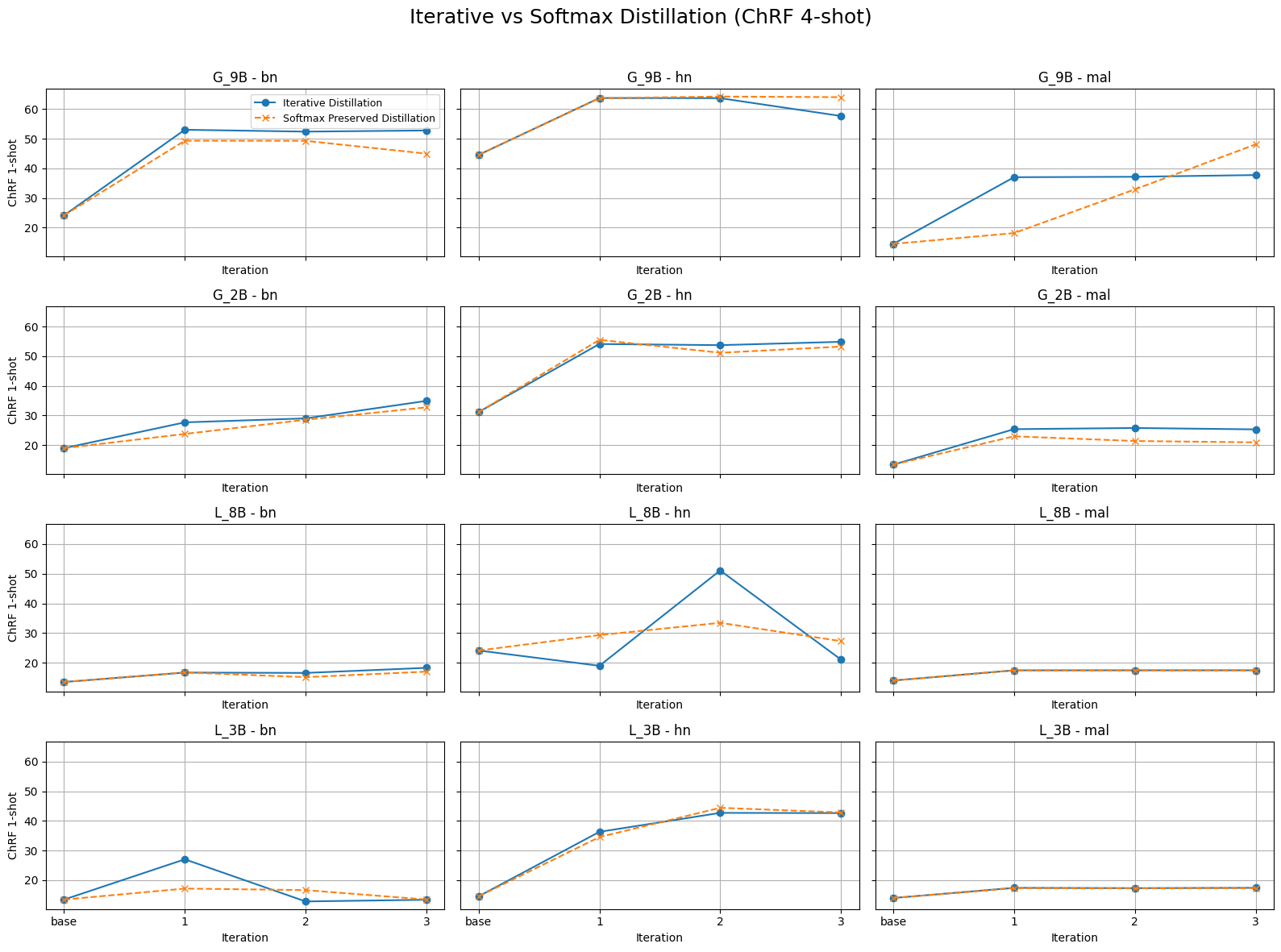}
    \caption{Comparison of the methods at 4-shot setting}
    \label{fig:vis_4shot}
\end{figure*}
\begin{figure*}
    \centering
    \includegraphics[width=0.9\linewidth]{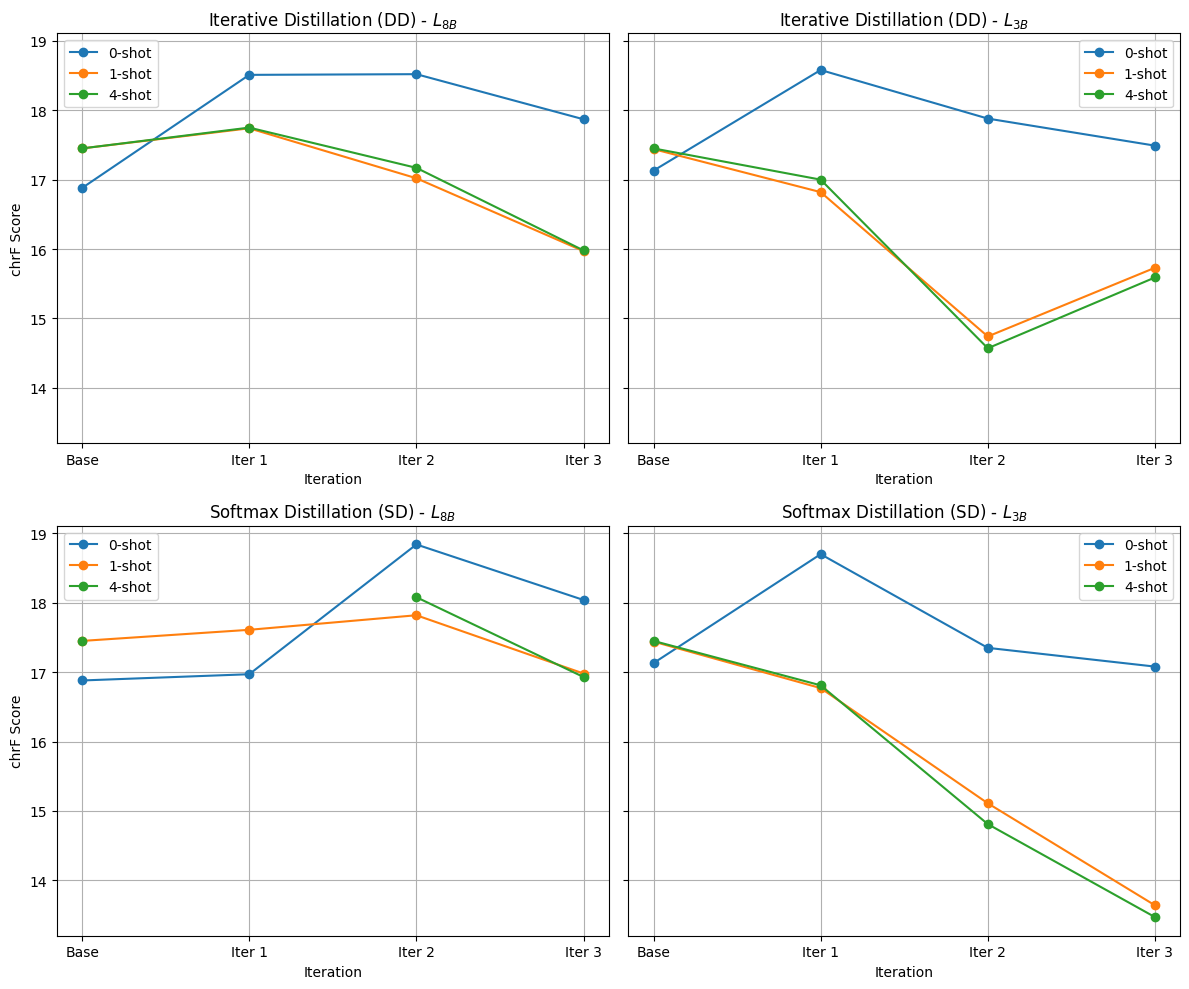}
    \caption{Comparison of the methods on the Manipuri}
    \label{fig:vis_manipuri}
\end{figure*}
\clearpage
\begin{figure*}
    \centering
    \includegraphics[width=\linewidth]{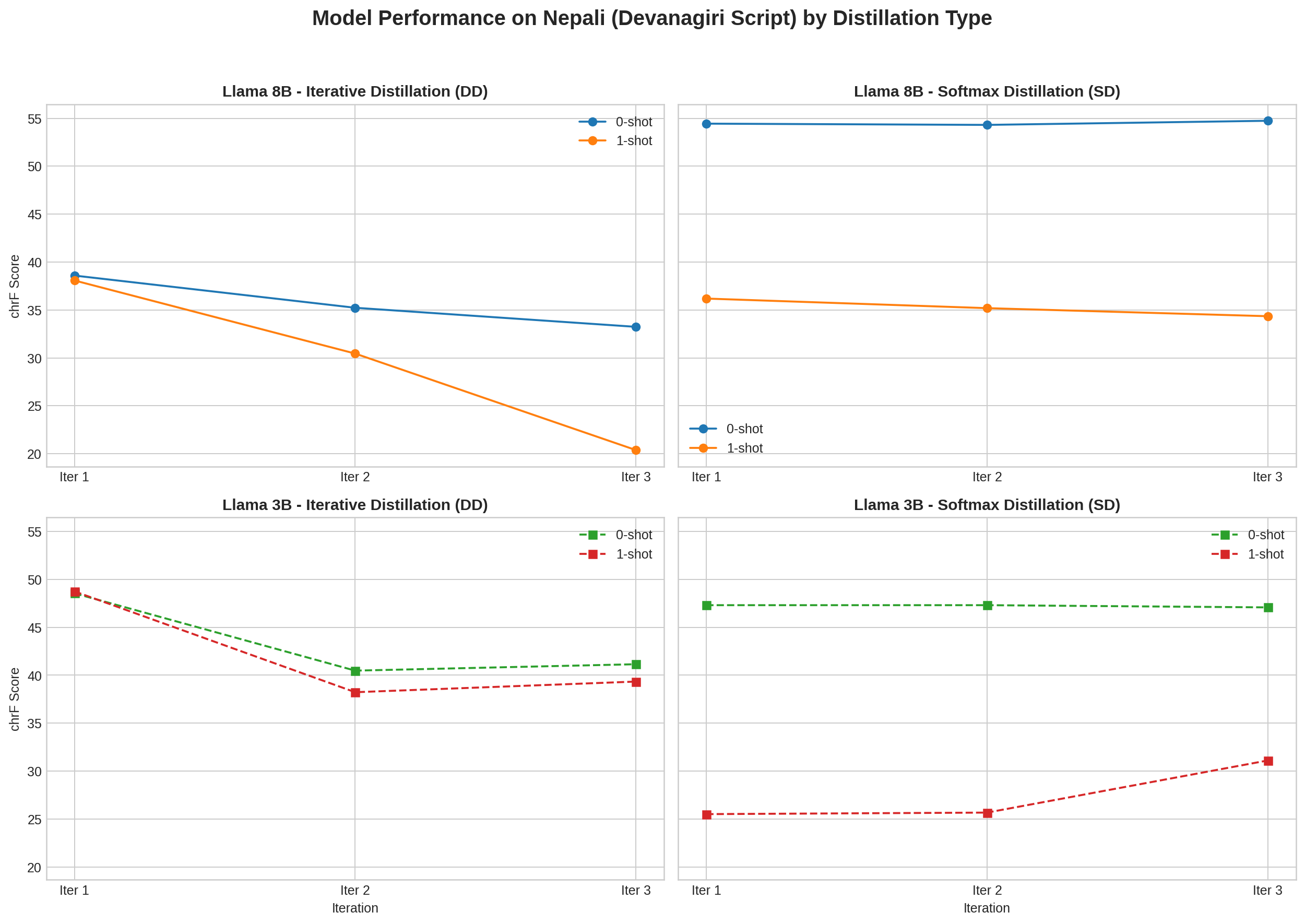}
    \caption{Comparison of the methods on the Nepali}
    \label{fig:vis_nepali}
\end{figure*}
\end{document}